\documentclass[review]{elsarticle}

\usepackage{lineno}
\usepackage{hyperref}
\usepackage{times}
\usepackage{soul}
\usepackage{url}
\usepackage[small]{caption}
\usepackage{graphicx}
\usepackage{amsmath}
\usepackage{amsthm}
\usepackage{amssymb}
\usepackage{bm}
\usepackage{booktabs}
\usepackage{algorithm}
\usepackage{algorithmic}
\usepackage{multirow}
\usepackage{latexsym}

\usepackage{tabularx}

\newcommand{\R}{\mathbb{R}}
\newcommand{\Gcal}{\mathcal{G}}
\newcommand{\Vcal}{\mathcal{V}}
\newcommand{\Ecal}{\mathcal{E}}
\newcommand{\Ocal}{\mathcal{O}}
\newcommand{\Ncal}{\mathcal{N}}

\DeclareMathOperator*{\softmax}{softmax}

\journal{Journal of \LaTeX\ Templates}

\begin{document}

\begin{frontmatter}

\title{Cross-Modal Knowledge Reasoning for Knowledge-based Visual Question Answering}

\author[mymainaddress,mysecondaryaddress]{Jing Yu \corref{mycorrespondingauthor}}
\cortext[mycorrespondingauthor]{Corresponding author}
\ead{yujing02@iie.ac.cn}
\author[mymainaddress,mysecondaryaddress]{Zihao Zhu}
\author[mythirdaddress]{Yujing Wang}
\author[myforthdaddress]{Weifeng Zhang}
\author[mymainaddress,mysecondaryaddress]{Yue Hu}
\author[mymainaddress,mysecondaryaddress]{Jianlong Tan}

\address[mymainaddress]{Institute of Information Engineering, Chinese Academy of Sciences, China}
\address[mysecondaryaddress]{School of Cyber Security, University of Chinese Academy of Sciences, China}
\address[mythirdaddress]{Microsoft Research Asia, China}
\address[myforthdaddress]{College of Mathematics, Physics and Information Engineering, Jiaxing University, China}

\begin{abstract}
Knowledge-based Visual Question Answering (KVQA) requires external knowledge beyond the visible content to answer questions about an image. This ability is challenging but indispensable to achieve general VQA. One limitation of existing KVQA solutions is that they jointly embed all kinds of information without fine-grained selection, which introduces unexpected noises for reasoning the correct answer. How to capture the question-oriented and information-complementary evidence remains a key challenge to solve the problem.
Inspired by the human cognition theory, in this paper, we depict an image by multiple knowledge graphs from the visual, semantic and factual views. Thereinto, the visual graph and semantic graph are regarded as image-conditioned instantiation of the factual graph.
On top of these new representations, we re-formulate Knowledge-based Visual Question Answering as a recurrent reasoning process for obtaining complementary evidence from multimodal information. To this end, we decompose the model into a series of memory-based reasoning steps, each performed by a \textbf{G}raph-based \textbf{R}ead, \textbf{U}pdate, and \textbf{C}ontrol (\textbf{GRUC}) module that conducts parallel reasoning over both visual and semantic information.  

By stacking the modules multiple times, our model performs transitive reasoning and obtains question-oriented concept representations under the constrain of different modalities. Finally, we perform graph neural networks to infer the global-optimal answer by jointly considering all the concepts. We achieve a new state-of-the-art performance on three popular benchmark datasets, including FVQA, Visual7W-KB and OK-VQA, and demonstrate the effectiveness and interpretability of our model with extensive experiments.
\end{abstract}


\end{frontmatter}

\section{Introduction}
\label{sec:intro}
\begin{figure}[tpb]
	\centering
	\setlength{\abovecaptionskip}{7pt}
	\includegraphics[width=0.7\columnwidth]{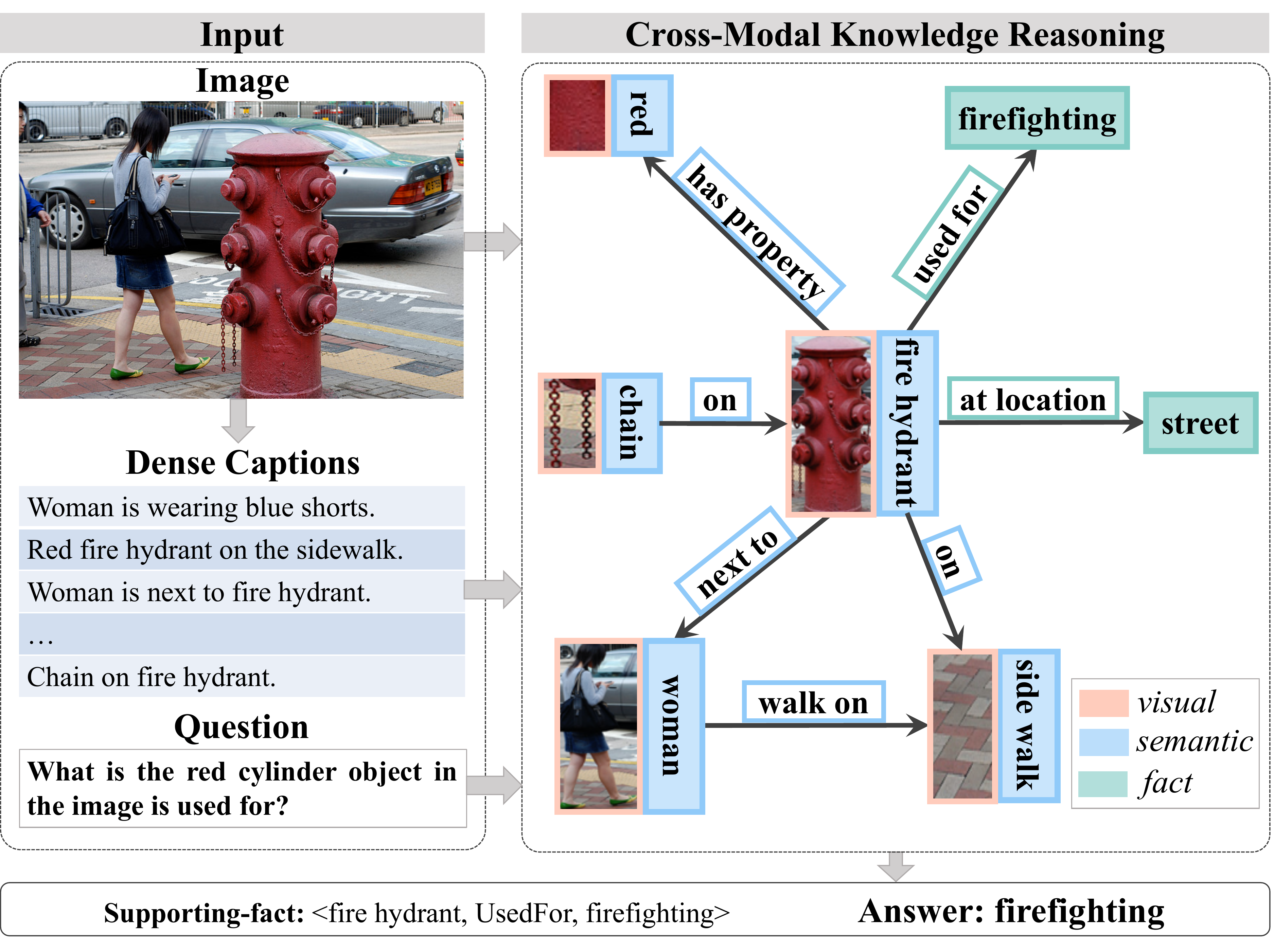}
	\caption{\small{An illustration of our motivation. We represent an image by graphs to associate visual, semantic and factual knowledge corresponding to the objects and relationships. Cross-modal knowledge reasoning is conducted on the graphs to infer the optimal answer.}}
	\label{fig:intro}
	\vspace{-2mm}
\end{figure}

Visual question answering (VQA) \cite{antol2015vqa} is an attractive research direction aiming to jointly analyze multimodal content from images and natural language. Equipped with the capacities of grounding, reasoning and translating, a VQA agent is expected to answer a question in natural language based on an image. Recent works~\cite{fang2019improving,yu2020reasoning} have achieved great success in VQA tasks that are answerable by solely referring to the visible content. However, such kinds of models are incapable of answering questions which require external knowledge beyond the visible content. 
Considering the question in Figure \ref{fig:intro}, the agent not only needs to visually localize `red cylinder', but also to semantically recognize it as `fire hydrant' and connects the knowledge that `fire hydrant is used for firefighting'. 
Therefore, how to collect question-oriented and information-complementary evidence from visual, semantic and knowledge perspectives is essential to achieve general VQA.

To advocate research in this direction, \cite{wang2018fvqa} introduces a Knowledge-based Visual Question Answering (KVQA) task, named as `Fact-based' VQA (FVQA), for answering questions by joint analysis of the image and the knowledge base of facts. The typical solutions for FVQA build a fact graph with fact triplets filtered by the visual concepts in the image and select one entity in the graph as the answer. 

Existing works \cite{wang2018fvqa,wang2017explicit} parse the question as keywords and retrieve the supporting-entity only by keyword matching. This kind of approaches is vulnerable when the question does not exactly mention the visual concepts (\textit{e.g.} synonyms and homographs) or the mentioned information is not captured in the fact graph (\textit{e.g.} the visual attribute `red' in Figure \ref{fig:intro} may be falsely omitted). To resolve these problems, \cite{narasimhan2018out} introduces visual information into the fact graph and infers the answer by implicit graph reasoning under the guidance of the question. However, they provide the whole visual information equally to each graph node by concatenation of the image, question and entity embeddings. Actually, only part of the visual content are relevant to the question and a certain entity.
Moreover, the fact graph here is still homogeneous since each node is represented by a fixed form of image-question-entity embedding, which limits the model's flexibility of adaptively capturing evidence from different modalities. A model has to be selective by choosing relevant information and avoiding unexpected noise.

\begin{figure}[tpb]
	\centering
	\setlength{\abovecaptionskip}{7pt}
	\includegraphics[width=0.7\columnwidth]{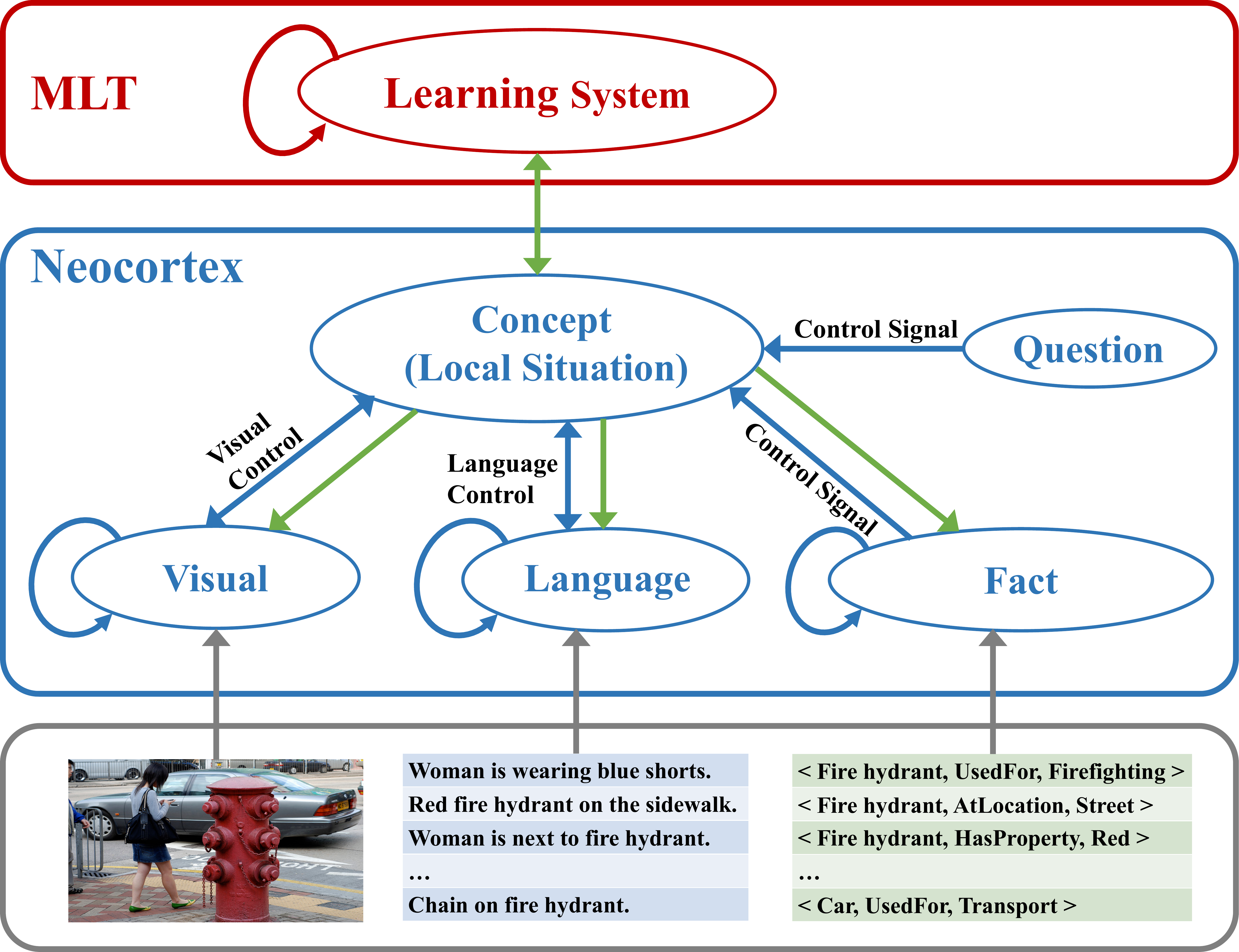}
	\caption{\small{The understanding system emulated by our model. The gray box contains information from multimodal sources. The blue box contains the neocortical sub-system, with each oval presenting an representation of a modal information. The blue arrows indicate learned connections that allow the representations to constrain each other.  The red box contains the medial temporal lobe (MLT) sub-system, which stores and processes all the representations from the neocortical system. The red arrow represents  self-connections that jointly consider all the representations for use. Green arrows connecting the red and blue ovals provide constraint between the two sub-systems.}}
	\label{fig:frameworkSimple}
	\vspace{-2mm}
\end{figure}

The recent proposed natural language understanding systems based on the cognition theory \cite{McClelland2019language} are consistent in that our brain is capable to adaptively combine multimodal input for understanding and reasoning. As proposed in \cite{McClelland2019language}, the understanding system contains two essential parts, where the neocortical sub-system (the blue box in Figure \ref{fig:frameworkSimple}) is responsible for selectively integrating linguistic and non-linguistic input to understand the object and situation while the medial temporal lobe (MTL) sub-system (the red box in  Figure \ref{fig:frameworkSimple}) aims to store and learn from the integrated embeddings of the neocortical sub-system states.  This understanding system is universal to tackle a wide range of natural language problems requiring external knowledge in multimodal format. In this perspective, KVQA problems can also be solved by this system by considering external knowledge in both linguistic (text and knowledge graph) and non-linguistic (image) format, which is flexible to choose task-relevant and content-complementary information for answer prediction.

Motivated by the proposed structure of understanding system in \cite{McClelland2019language}, we first introduce a novel scheme to depict an image by unifying the graph representation of different modalities, including the visual graph, semantic graph and fact graph. Specifically, the object appearance and their relationships are kept in the visual graph, the high-level abstraction is provided in the semantic graph and the corresponding factual knowledge  is supported in the fact graph, which imitates  distinct areas in neocortical sub-system  that processes  each  input modality (the three blue ovals at bottom of neocortex). Graph representation is suitable for modeling the objects (entities) and their relationships of input and beneficial for connecting different modalities to constrain each other. Then we integrate cross-modal knowledge corresponding to the same concept (the blue oval at top of neocortex) by a series of memory-based reasoning steps. In order to select complementary knowledge from different modalities for integration, we propose a constraint satisfaction process in which the information in one modality influences the selection of information in another modality. To this end, we perform each reasoning step  by a \textbf{G}raph-based \textbf{R}ead, \textbf{U}pdate, and \textbf{C}ontrol (\textbf{GRUC}) module that conducts parallel reasoning over both visual and semantic information: the control unit updates the control signal for extracting a knowledge vector from the knowledge graphs (visual and semantic); the read unit generates the knowledge vector from the knowledge graphs upon the constrain of the control signal; the update unit integrates the knowledge vector into the control signal as well as the knowledge graph for memory update. After multiple reasoning steps, we obtain complementary evidence from different modalities and fuse them adaptively to reason about the global-optimal answer via a graph neural network, which can be seen as the learning system in MLT. The main contributions can be summarized as follows:

(1) We novelly depict multimodal knowledge sources by multiple knowledge graphs from the visual, semantic and factual views, which unifies the representations of different modalities in graph domain and thus benefits for structure preserving and relational reasoning. %We consider these three modalities jointly and achieve significant improvement over state-of-the-art solutions. 
Thereinto, introducing the semantic graph for high-level abstraction brings remarkable improvement in KVQA, which has been less studied in previous work. %enrich the representations of objects and relationships; 

(2) We propose a recurrent reasoning model that has three obvious novelties: First, it is a parallel reasoning model that applies modality-oriented controllers for reasoning over different modalities in a parallel mode, which can be easily extended to involve more modalities; Second, our model is designed for reasoning upon graph-structured multimodal data, aiming to consider the essential structural information in the reasoning process; Third, the basic reasoning module GRUC is a modular architecture consisting of Read, Update and Control units, which supports more explicit and structured reasoning.

(3) The proposed model remarkably outperforms  state-of-the-art approaches on three benchmark datasets, including FVQA, Visual7W-KB and OK-VQA, which demonstrates the feasibility and effectiveness of the proposed model. Through ablative studies, we prove how each of the proposed components contributes to the improvement.

(4) The proposed model has good interpretability. It automatically tells which concept (entity) and modality (visual, semantic or factual) have more contributions to answer the question through visualization of attention weights in GRUC and gate values in the fusion process. Meanwhile, the model can also reveal the knowledge selection mode from different modalities according to the complexity of the questions.

\section{Related Work}
\label{sec:relatedWork}

\subsection{Visual Question Answering}
The Visual Question Answering (VQA) task requires the agent to answer a  question in natural language according to the visual content in an image, which demands for  comprehending and reasoning about both visual and textual information. 
The typical solutions for VQA are based on the CNN-RNN architecture \cite{ren2015exploring} %,malinowski2015ask
that coarsely fuses the global visual and textual features as clues to predict the answer. For better combining textual information with visual information, bilinear pooling approaches \cite{yu2017multi,ben2019block} have been proposed to fuse multimodal features in fine-grained mode. However, the above approaches leverage all the information in the image and question, which may introduce redundant or noisy information to the prediction stage.
%The current dominant trend mostly adopt the pipeline consisting of image encoder, question encoder, multimodal fusion and answer predictor~\cite{wu2017visual}.
%Early methods~\cite{ren2015exploring,malinowski2015ask} directly using global visual features extracted from CNN
To alleviate this problem, various attention mechanisms have been exploited in VQA tasks \cite{lu2016hierarchical,wang2020long}  to highlight visual objects that are relevant to the question. 
%~\cite{nam2017dual} proposed to perform question-guided image attention and image-guided question attention collaboratively. ~\cite{yang2016stacked} proposed a stacked attention network to learn the attention iteratively.
~\cite{anderson2018bottom} introduced a bottom-up and top-down attention mechanism to learn the attention on candidate objects rather than spatial grids.
However, they treat objects in an image independently and ignore their informative relationships.

Humans’ ability of combinatorial generalization highly depends on the mechanisms of reasoning over relationships. 
%\textit{or ``Relationship information is essential for improving modern AI’s capacity for combinatorial generalization and reasoning.''}
Consistent with such idea, there is an emerging trend to depict objects and visual relationships in an image by graph structure to  support reasoning in VQA.  %\cite{norcliffe2018learning} and other vision-language tasks \cite{jiang2019dualvd,wang2020learning}.
One kind of approaches performs one-step relational reasoning to infer the answer. \cite{Santoro2017simple}  proposed a Relation Network (RN) to model all the implicit relationships among objects in the image by multi-layer perceptrons (MLPs). Then, the relationships are summed and fed into other MLPs to predict the answer. This approach brings high computational cost and makes it hard to perform multi-step reasoning. \cite{norcliffe2018learning} refined the relationships by a ranking strategy and lowered the computational complexity. However, a large number of compound questions require  multi-step reasoning. To this end, \cite{wu2018chain} proposed multi-step attention to reason over both original objects and new compound objects and infer the answer progressively. Additionally, \cite{hudson2019learning} transformed both the visual and textual moldalities into concept-based graph representations and performed sequential reasoning over the graph by the neural state machine. Furthermore, \cite{jiang2019dualvd} exploited semantic captions to further enrich the graph-based representations for multi-step reasoning. Reasoning approaches in the above work are always on visual and textual features, which cannot be extended to involve external knowledge.
To go one step further, our model pays attention to not only original input features but also external knowledge 
during progressive reasoning.

\subsection{Incorporating External Knowledge in VQA}
Human easily combine visual observation with external knowledge for answering questions, which remains challenging for algorithms.
To bridge this discrepancy, \cite{wang2018fvqa} introduced a Fact-based VQA (FVQA) task, which additionally provides a knowledge base of facts and associates each question with a supporting-fact.
Recent works based on FVQA generally select one entity from fact graph as the answer and falls into two categories: query-mapping based methods and learning based methods.
On the one hand, \cite{wang2017explicit} reduced the question to one of the available query templates and this limits the types of questions that can be asked. 
\cite{wang2018fvqa} automatically classified and mapped the question to a query which does not suffer this constraint.
%which runs over the combined image and knowledge information. 
Among both methods, however, visual information is used to extract facts but not introduced during the reasoning process.
%they are vulnerable to misconceptions caused by synonyms and homographs. 
On the other hand, \cite{narasimhan2018straight} learned a similarity score between the representations of fact and image-question pair.
~\cite{narasimhan2018out} applied graph convolutional networks on the fact graph where each node is represented by the fixed form of image-question-entity embedding. However, the visual information is wholly provided which may introduce redundant information for reasoning the answer. The same problem also exists in \cite{li2017incorporating}, although they leveraged dynamic memory network instead of graph convolutional netowrk to incorporate the external knowledge. Recent work \cite{marino2019ok} proposed a new knowledge-based task OK-VQA and introduced a retrieval-based model to extract the correct answer from Wikipedia. 
Different from previous work, in this paper, we decipt an image by multimodal kwnoledge graphs and perform cross-modal  reasoning via a memory-based recurrent network to capture complementary evidence from different modalities.

\subsection{Graph Neural Networks}
The core module GRUC in our proposed model is a novel graph-based neural network. In this subsection, we briefly review related Graph Neural Networks (GNNs) and highlight differences between previous work and ours. Approaches based on GNNs \cite{scarselli2008graph} repeatedly perform a message passing process over the graph by aggregating and updating information between nodes. Relying on spectral graph theory, \cite{kipf2016semi} exploited simplified Chebyshev polynomials to construct localized polynomial filters for graph convolution in graph convolutional networks (GCN). Attention mechanisms have been introduced in \cite{velivckovic2017graph} to learn the weights over edges for convolution operations. Lanczos-based method \cite{liao2018lanczosnet} has been explored for graph convolution for the purpose of acceleration. Our model is closely related to the Gated Graph Sequence Neural Networks (GGS-NN) \cite{li2016gated} which updates GNNs by adding gated recurrent unit. Different from GGS-NN that both convolution operation and recurrent propagation are performed in the same graph, our GRUC module aggregates information from external knowledge graph for node updating and recurrent propagation in another fact graph.  
Our model is also related to the heterogeneous graph neural networks since the model is reasoning over multimodal graphs. \cite{schlichtkrull2018modeling} generalized graph convolutional network to handle different relationships between entities in a knowledge base, where edges with distinct relationships are encoded independently.
\cite{wang2019heterogeneous} proposed heterogeneous graph attention networks with dual-level attention mechanism.
All the above approaches model different types of nodes and edges in an unified graph.
In contrast, the heterogeneous graph in this work contains multiple layers of subgraphs and each layer consists of nodes and edges coming from different modalities. For this specific constrain, we propose the parallel reasoning model that applies modality-oriented controllers for reasoning over different modalities in a parallel mode.

\section{Methodology}
\label{sec:methodology}
\setlength{\abovecaptionskip}{7pt}
\begin{figure*}[t]
	\includegraphics[width =\textwidth]{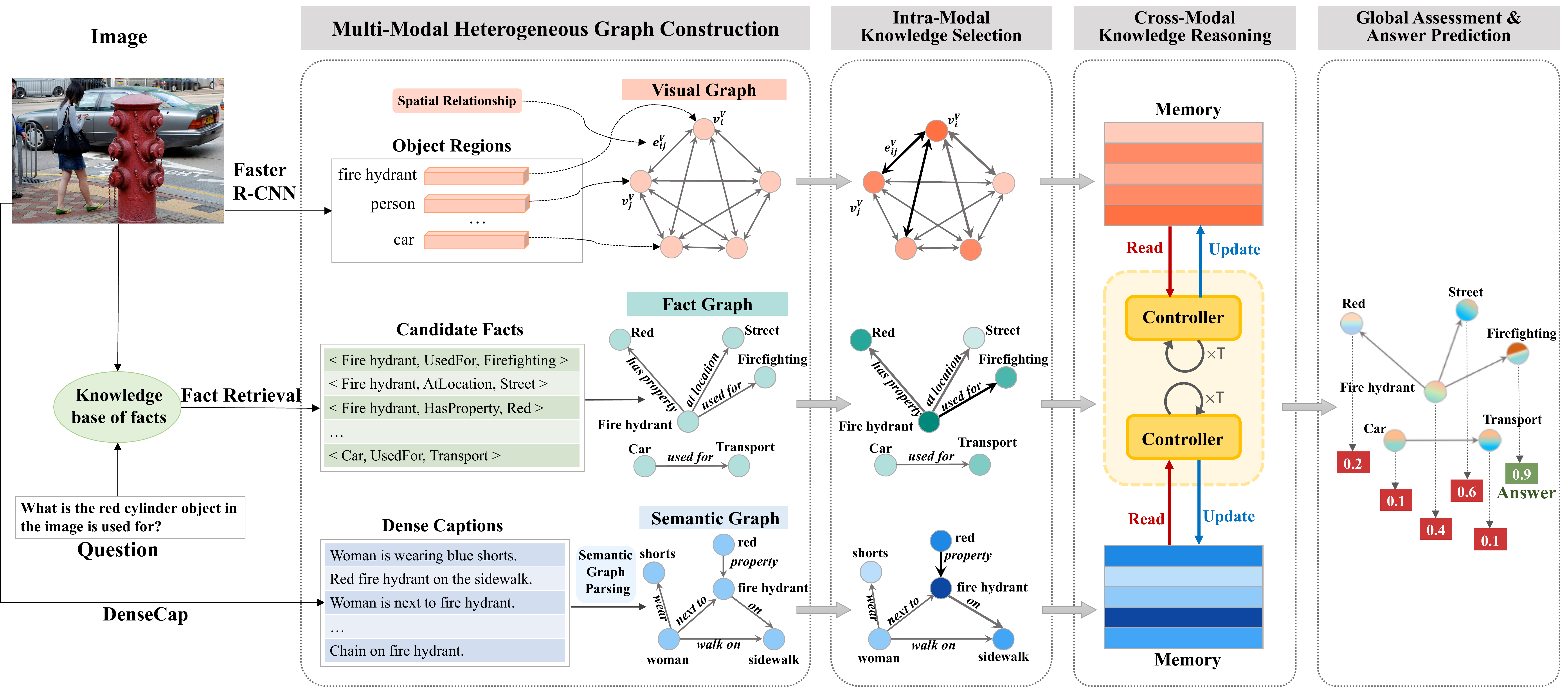}
	\caption{\small{An overview of our model. The model contains four parts: multimodal heterogeneous graph construction,
	intra-Modal knowlegdge selection, cross-modal knowledge reasoning, and global assessment and answer prediction.}
	}
	\label{fig:model}
	\vspace{-2mm}
\end{figure*} 

Given an image $I$ and a question $Q$, the task aims to predict an answer $A$ by leveraging the external knowledge. In this work, we focus on external knowledge in the form of knowledge graph, which consists of a set of triplet facts , \textit{i.e.} $<e_1, r, e_2>$, where $e_1$ is a visual concept in the image, $e_2$ is an attribute or phrase and $r$ represents the relationship between $e_1$ and $e_2$. %Each question $Q$ is associated with a ground truth supporting fact. 
%To answer the question, 
The key is to choose a correct concept, \textit{i.e.} either $e_1$ or $e_2$, from the supporting fact as the predicted answer. 

The proposed model mainly contains four parts: (1) Firstly,  \textit{multimodal Knowledge Graph Construction} (Section \ref{subsec:graphConstruction}) represents knowledge from different modalities by different knowledge graphs, including the visual graph, semantic graph and fact graph, imitating distinct brain areas that represent each input modality; (2) Then,
\textit{Intra-Modal Knowledge Selection} (Section \ref{subsec:IntraConv}) selects question-oriented knowledge from each modality of knowledge graph by intra-modal graph convolution; (3) Afterwards,  \textit{Cross-Modal Knowledge Reasoning} (Section \ref{subsec:crossReasoning}) performed by the GRUC Network iteratively gathers complementary evidence from the visual and semantic knowledge graphs under the guidance of the question and the facts. In the end of reasoning steps, we fuse the evidence from three modalities to obtain the representation of each concept. %Both of above convolutions are guided by question based on the node attention as well as the edge attention. 
(4) Finally, \textit{Global Assessment and Answer Prediction} (Section \ref{subsec:globalAnswer}) aims to jointly analyze all the concepts via graph convolutional networks and predict the optimal answer by a binary classifier. Figure \ref{fig:model} gives detailed illustration of our model.

\subsection{Multimodal Graph Construction}
\label{subsec:graphConstruction}

\subsubsection{Visual Graph Construction}
Since most of the questions in KVQA grounded in the visual objects and their relationships, we construct a fully-connected visual graph to represent such evidence at appearance level. Given an image $I$, we use Faster-RCNN \cite{ren2015faster} to identify a set of objects $\Ocal = \{o_i\}_{i=1}^K$ ($K$ = 36), where each object $o_i$ is associated with a visual feature vector $\bm{v}_i \in \R^{d_v}$ ($d_v$ = 2048), a bounding-box feature vector $\bm{b}_i\in \R^{d_b}$ ($d_b$ = 4) and a corresponding label. 
Specifically, $\bm{b}_i = [x_i, y_i, w_i, h_i]$, where $(x_i, y_i)$, $h_i$ and $w_i$ respectively denote the coordinate of the top-left corner, the height and width of the bounding box.
We construct a visual graph $\Gcal^V=(\Vcal^V,\Ecal^V)$ over $\Ocal$, where $\Vcal^V =\{v_i^V\}_{i=1}^K$ is the node set and each node $v^V_{i}$ corresponds to a detected object $o_i$. 
The feature of node $v^{V}_{i}$ is represented by $\bm{v}^V_i$. Each edge $e^V_{ij} \in \Ecal^V$ denotes the relative spatial relationships between two objects. We encode the edge feature by a 5-dimensional vector, \textit{i.e.} $\bm{r}^V_{ij} = [\frac{x_j-x_i}{w_i},\frac{y_j-y_i}{h_i},\frac{w_j}{w_i},\frac{h_j}{h_i},\frac{w_jh_j}{w_ih_i}]$.
%Thus, graph $\Gcal^V$ contains appearance-level visual information of the image.

\subsubsection{Semantic Graph Construction}
In addition to visual information, high-level abstraction of the objects and relationships by natural language provides essential semantic information. Such abstraction is indispensable to associate the visual objects in the image with the concepts mentioned in both questions and facts. In our work, we leverage dense captions \cite{johnson2016densecap} to extract a set of local-level semantics in an image, ranging from the properties of a single object (color, shape, emotion, $etc.$) to the relationships between objects (action, spatial positions, comparison, $etc.$). We decipt an image by $D$ dense captions, denoted as $Z=\{z_i\}_{i=1}^D$, where $z_i$ is a natural language description about a local region in the image. Instead of using monolithic embeddings to represent the captions, we exploit modelling them by a graph representation, denoted as $\Gcal^S=(\Vcal^S,\Ecal^S)$, which is constructed by a semantic graph parsing model \cite{anderson2016spice}. %We convert the dense captions into a semantic graph, denoted as $\Gcal^S(Z)=(\Vcal^S,\Ecal^S)$, where 
The node $v^S_i \in \Vcal^S$ represents the name or attribute of an object extracted from the captions while the edge $e_{ij}^S\in \Ecal^S$ represents the relationship between $v^S_i$ and $v^S_j$. We use the averaged GloVe embeddings to represent $v^S_i$ and $e_{ij}^S$, denoted as $\bm{v}_{i}^S$ and $\bm{r}_{ij}^S$, respectively. The graph representation retains the relational information among concepts and unifies the representations in graph domain, which is better for explicit reasoning across modalities.

\subsubsection{Fact Graph Construction}

%Firstly candidate facts are retrieved for graph construction. 
To find the optimal supporting-fact, we first retrieve relevant candidate facts from knowledge base of facts following a score-based approach \cite{narasimhan2018out}. We compute the cosine similarity of the GloVe embeddings of every word in the fact with the words in the question and the words of visual concepts detected in the image. Then we average these values to assign a similarity score to the fact. The facts are sorted based on the similarity and the 100 highest scoring facts are retained, denoted as $f_{100}$. A relation type classifier is trained additionally to further filter the retrieved facts.
Specifically, we feed the last hidden state of LSTM to an MLP layer to predict the relation type $\hat{r}_i$ of a question. We retain the facts among $f_{100}$ only if their relationships agree with $\hat{r}_i$, \textit{i.e.} $f_{rel}=f\in f_{100}:r(f) \in\{\hat{r}_i\}$ ($\{\hat{r}_i\}$ contains top-3 predicted relationships in experiments). 
Then a fact graph $\Gcal^F=(\Vcal^F,\Ecal^F)$ is built upon $f_{rel}$ as the candidate facts can be naturally organized as graphical structure.
Each node $v_i^F \in \Vcal^F$ denotes an entity in $f_{rel}$ and is represented by GloVe embedding of the entity, denoted as $\bm{v}_i^F$.
Each edge $e_{ij}^F \in \Ecal^F$ denotes the relationship between $v_i^F$ and $v_j^F$ and is represented by GloVe embedding $\bm{r}_{ij}$. The topological structure among facts can be effectively exploited by jointly considering all the entities in the fact graph.

\subsection{Intra-Modal Knowledge Selection}
\label{subsec:IntraConv}
Since each layer of graphs contains modality-specific knowledge relevant to the question, we first select valuable evidence independently from the visual graph, semantic graph and fact graph by \textit{Visual-to-Visual Convolution}, \textit{Semantic-to-Semantic Convolution} and \textit{Fact-to-Fact Convolution}, respectively. These three convolutions share the common operations but differ in their node and edge representations corresponding to the graph layers. Thus we omit the superscript of node representation $\bm{v}$ and edge representation $\bm{r}$ in the rest of this section. 
We first perform attention operations to highlight the nodes and edges that are most relevant to the question $q$ and consequently update node representations via intra-modal graph convolution. %Considering this process is identical to $\Gcal^{V}$, $\Gcal^{S}$ and $\Gcal^{F}$, we omit the superscript of $\Gcal$ in the rest of this section. 
This process mainly consists of the following three steps:

\textbf{Question-guided Node Attention.}
We first evaluate the relevance of each node corresponding to the question by attention mechanism. The attention weight for $v_i$ is computed as:
\begin{equation}\label{eq:node attention}
\alpha_i=\softmax(\bm{w}^T_a\tanh({\textbf{W}_{1} }\bm{v}_i + \textbf{W}_{2}\bm{q}))
\end{equation}
where $\textbf{W}_{1}$,$\textbf{W}_{2}$ and $\bm{w}_a$ (as well as $\textbf{W}_{3}$,..., $\textbf{W}_{12}$, $\bm{w}_b$, $\bm{w}_c$ mentioned below) are learned parameters. $\bm{q}$ is question embedding encoded by the last hidden state of LSTM.

\textbf{Question-guided Edge Attention.} Under the guidance of question, we then evaluate the importance of edge $e_{ji}$ constrained by the neighbor node $v_j$ regarding to $v_i$ as follows:
\begin{equation}\label{edge attention}
\beta_{ji}=\softmax(\bm{w}^T_b\tanh({\textbf{W}_{3} }\bm{v'}_{j} + \textbf{W}_{4}\bm{q'}))
\end{equation}

where $\bm{v'}_{j}=\textbf{W}_5[\bm{v}_j,\bm{r}_{ji}]$, $\bm{q'}=\textbf{W}_{6}[\bm{v}_i,\bm{q}]$ and $[\cdot,\cdot]$ denotes concatenation operation.% and $\Ncal_i$ denotes neighbourhood of node $v_i$.

\textbf{Intra-Modal Graph Convolution.}
Given the node and edge attention weights learned in Eq. {\ref{eq:node attention}} and Eq. \ref{edge attention}, the node representations of each layer of graphs are updated following the message-passing framework~\cite{gilmer2017neural}. We gather the neighborhood information and update the representation of $v_i$ as follows:
\begin{align}
&\bm{m}_i=\sum_{j\in\mathcal{N}_i}\beta_{ji}\bm{v'}_{j} \\
&\hat{\bm{v}}_i=\text{ReLU}(\textbf{W}_7[\bm{m}_i,\alpha_i\bm{v}_i])\label{eq:4}
\end{align}
where $\Ncal_i$ is the neighborhood set of node $v_i$.
%$\alpha_i$ and $\beta_{ij}$ are question-guided node and edge attention value computed in the previous section.
We conduct the above intra-modal knowledge selection on $\Gcal^V$, $\Gcal^S$ and $\Gcal^F$ independently and obtain the updated node representations, denoted as $\{\hat{\bm{v}}_i^V\}_{i=1}^{\Ncal^V}$, $\{\hat{\bm{v}}_i^S\}_{i=1}^{\Ncal^S}$ and $\{\hat{\bm{v}}_i^F\}_{i=1}^{\Ncal^F}$ accordingly. %We use $\text{GNN}(\Gcal,\bm{q})$ to denote oprations in Eq. \ref{eq:node attention} to Eq. \ref{eq:4} below.

\begin{figure}[t]
	\centering
	\setlength{\abovecaptionskip}{7pt}
	\includegraphics[width=\columnwidth]{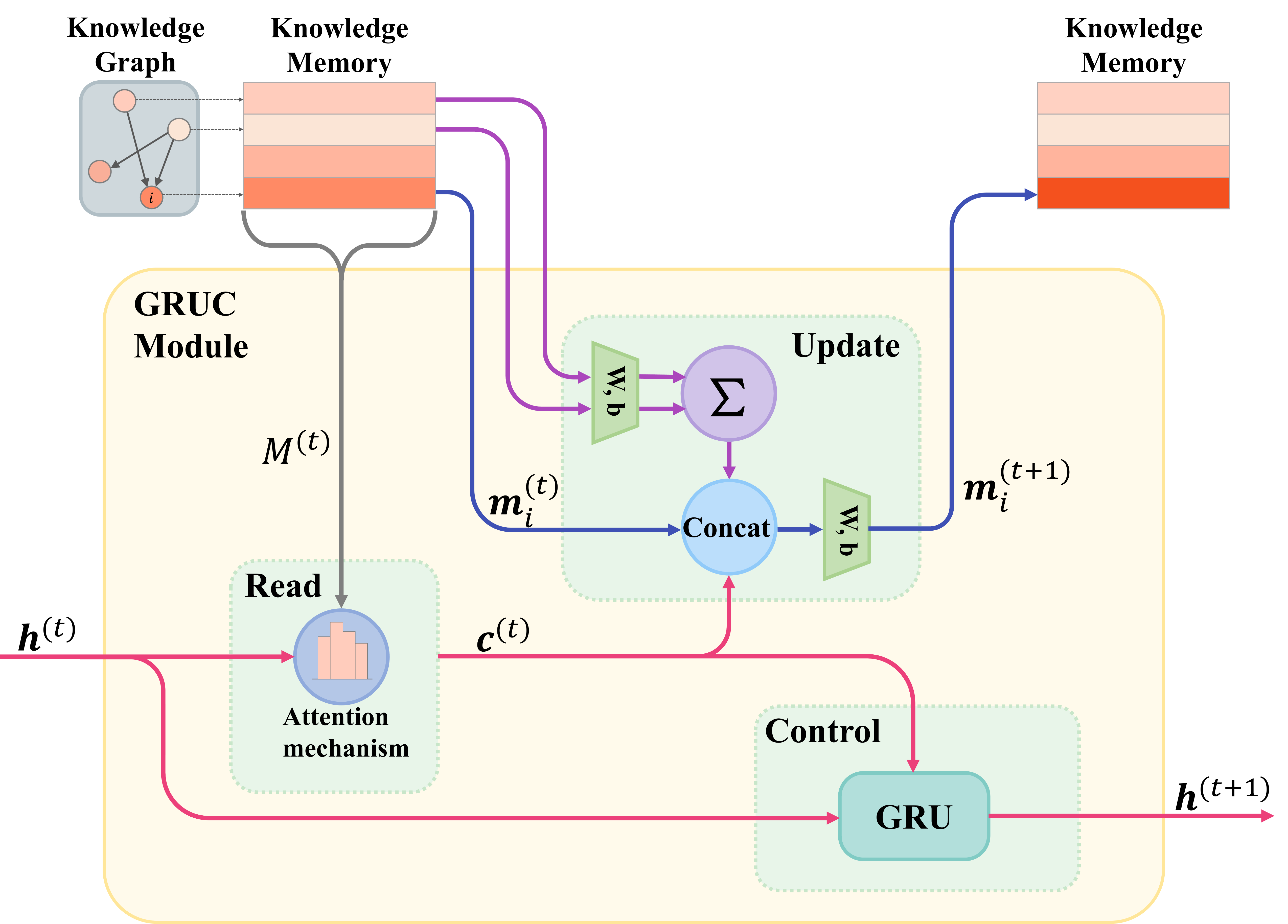}
	\caption{\small{The GRUC module architecture.}}
	\label{fig:GRUC}
	\vspace{-2mm}
\end{figure}

\begin{figure}[t]
\setlength{\abovecaptionskip}{7pt}
	\includegraphics[width =\columnwidth]{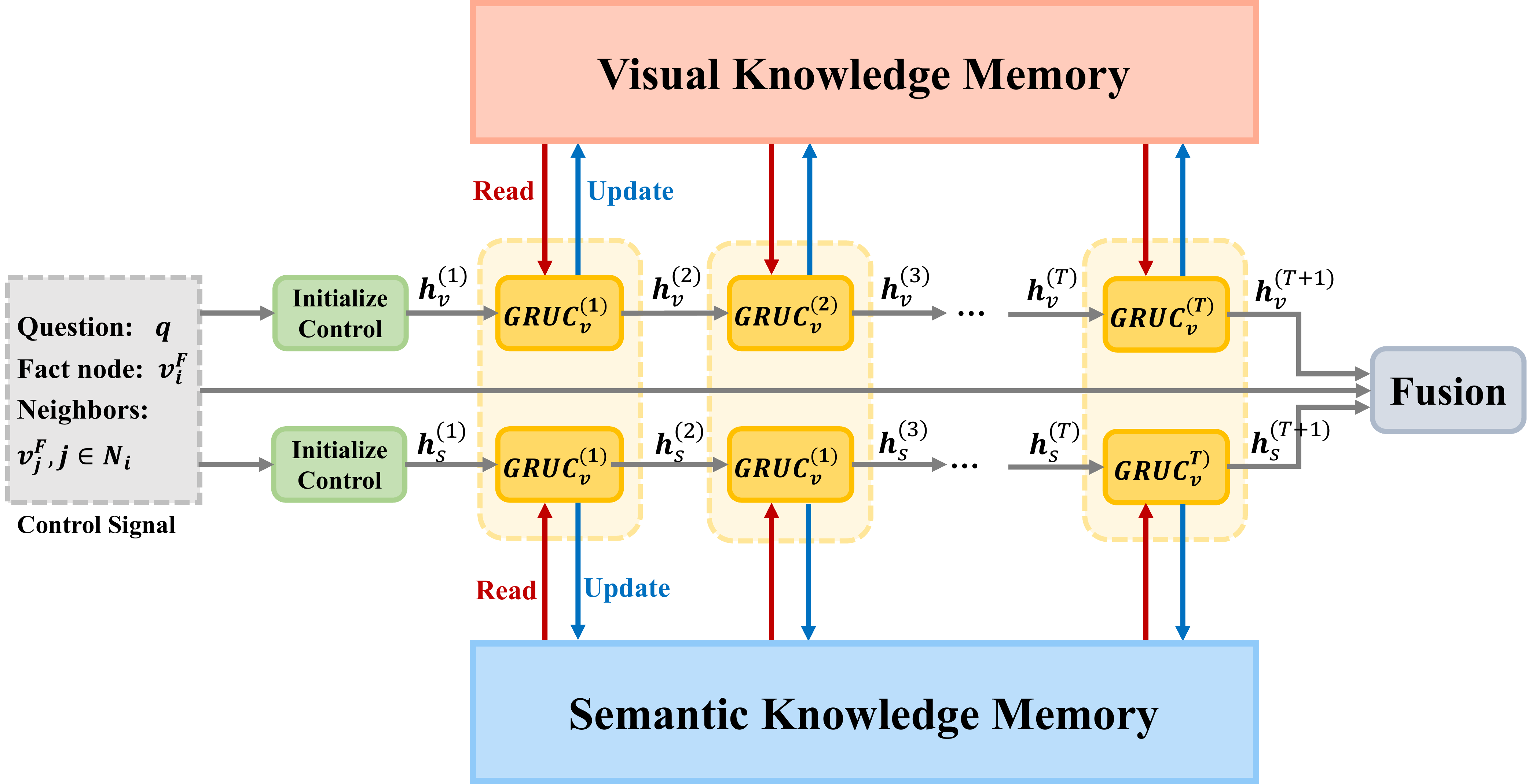}
	\centering
	\caption{\small{An illustration of the GRUC network.}}
	\label{fig:GRUCnetwork}
	\vspace{-2mm}
\end{figure}

\subsection{Cross-Modal Knowledge Reasoning}
\label{subsec:crossReasoning}

The key of cross-modal knowledge graph reasoning module is to reason on the fact graph by jointly considering all candidate facts. However, the entities in the fact graph provide insufficient knowledge to reason about the globally optimal answer which need to be complemented with correlated knowledge from other modalities. The process is performed by our proposed GRUC Network, a memory-based reasoning architecture by sequencing a recurrent Graph-based Read, Update and Control (GRUC) module. After multi-step reasoning, we fuse the multimodal knowledge for each entity and achieve more comprehensive understanding of an entity, which we rename it as `concept' afterwards. The GRUC Network contains two components: a recurrent GRUC module and a multimodal feature fusion module, as introduced below.

\subsubsection{The Recurrent GRUC Module}

%The proposed GRUC network is a multi-step reasoning process by stacking GRUC modules $T$ times. 
The GRUC module aims to gather question-oriented knowledge from different modalities corresponding to the same concept.
However, it's non-trivial to align the knowledge from different modalities to the same concept since there rarely exists explicit one-to-one mappings across modalities. Therefore, we propose to gather concept-relevant knowledge in an implicit way. Since the answer comes from one entity in the fact graph in this task, we regard each entity in the fact graph as  a concept and gather complementary knowledge from the visual graph and the semantic graph to this concept in the fact graph in the GRUC module.
In this way, the recurrent GRUC modules can be performed over all the concepts in the fact graph in parallel and obtaining the relevant visual knowledge, semantic knowledge and fact knowledge for each concept as the output. 

Figure \ref{fig:GRUCnetwork} shows the GRUC network operated on the concept (i.e. fact node) $v_i^F \in \Vcal^F$. The recurrent GRUC modules (yellow boxes) iteratively gather knowledge from the visual knowledge memory (red box) and the semantic memory (blue box) in a parallel way. For each step $t$ ($t=1,2,...,T$ ) in the reasoning process, the $t^{th}$ GRUC module maintains two hidden states: visual control state $\bm{h}^{(t)}_v$ and semantic control state $\bm{h}^{(t)}_s$, initialized by the question $q$, $v_i^F$ and its neighborhood nodes (gray and green boxes) to learn $\bm{h}^{(1)}_v$ and $\bm{h}^{(1)}_s$, respectively. Since the reasoning processes for the visual part and the semantic part share thee common operations but differ in the graph representations, we omit the subscript of hidden state representation $\bm{h}^{(t)}$ in the rest of this section. Figure \ref{fig:GRUC} shows the architecture of the GRUC module, which consists of the Control Unit, Read Unit and Update Unit.

\textbf{The Control Unit.} %For each node $v_i$ of the fact graph, we denote the hidden state of RNN-based controller at the step $t$ as $\bm{h}^{(t)}$.
The control unit determines the \textit{control state} $\bm{h}^{(t)}$ that guides the module to adaptively select complementary knowledge from the knowledge memory for obtaining a more comprehensive concept representation.
At the first reasoning step, the control unit is initialized by fusing the question representation $\bm{q}$, the entity representation $\hat{\bm{v}}_i^F$ and its neighborhood information as:
\begin{align}
\label{eq:controInit}
&\bm{h}^{(1)}=\textbf{W}_8[\bm{q},\hat{\bm{v}}_i^F,\bm{c}^F_i]\\
&\bm{c}^F_i=\sum_{j\in \Ncal_i}\hat{\bm{v}}_j^F
\end{align}
where $\Ncal_i$ represents a set of 1-hop neighboring nodes regarding the entity $v_i^F$. In the $t^{th}$ reasoning step, the control state will be updated with the contextual vector $\bm{c}^{(t)}$ (will be introduced in the Read Unit) extracted from the knowledge memory and previous hidden state $\bm{h}^{(t)}$ via Gated Recurrent Unit (GRU) \cite{cho2014learning}, the update operation is defined as follows:
\begin{equation}
\bm{h}^{(t+1)}=GRU(\bm{h}^{(t)},\bm{c}^{(t)})
\end{equation}

Then updated control state $\bm{h}^{(t+1)}$ is used to control the reasoning process in the next step.

\textbf{The Read Unit.} In the $t^{th}$ reasoning step, the read unit gathers the required knowledge $\bm{c}^{(t)}$ in the knowledge base (visual graph or semantic graph) under the guidance of the control state $\bm{h}^{(t)}$ and previous knowledge memory $\bm{M}^{(t)}$$=$$\{$$\bm{m}_{1}^{(t)}$, $\bm{m}_{2}^{(t)}$,$\dots$, $\bm{m}_{N}^{(t)}$$\}$ ($N$ is the number of memory entries). Specifically,  $\bm{M}^{(t)}$ is a graph-structured memory, where each entry represents the node in the knowledge graph ($\Gcal^V$ or $\Gcal^S$) and the relationships between entries are the corresponding edges in the knowledge graph. The initial representation of each entry $\bm{m}_{j}^{(1)}$ is the corresponding node representation obtained after intra-modal knowledge selection, \emph{i.e.} $\bm{m}_{j}^{(1)}=\hat{\bm{v}}_j$.

We compute the required knowledge $\bm{c}^{(t)}$  via an attention component. The attention value $\gamma_{j}^{(t)}$ for the memory entry  $\bm{m}_{j}^{(t)}$ is calculated under the guidance of the control state $\bm{h}^{(t)}$ as:
\begin{align}
 &\bm{a}_{j}^{(t)}=\tanh(\textbf{W}_9\bm{h}^{(t)}+\textbf{W}_{10}\bm{m}_j^{(t)})\\
 &\gamma_{j}^{(t)}=\softmax(\bm{w}_c^T\bm{a}_{j}^{(t)}) \label{mem_att}
 %&\gamma_{j}^{(t)}=\frac{\exp(e_{j}^{(t)})}{\sum_{k=1}^{|M|}\exp{ e_{k}^{(t)} }}
\end{align}

Then we generate the required knowledge $\bm{c}^{(t)}$ by weighting over all the memory entries defined as:
\begin{equation}
\bm{c}^{(t)}=\sum_{j=1}^{N}\gamma_{j}^{(t)}\bm{m}_j^{(t)}
\end{equation}

The function of $\bm{c}^{(t)}$ is divided into two folds: On the one hand, it is used to update the control unit and generate the control state $\bm{h}^{(t+1)}$ for the next reasoning step; On the other hand, $\bm{c}^{(t)}$ records the required knowledge retrieved from the knowledge graph during the transitive reasoning process and will be regarded as the knowledge representation of corresponding modality for the final multimodal feature fusion. 

\textbf{The Update Unit.} Unlike the static knowledge representations, our proposed knowledge memory will be updated adaptively during each reasoning step via the Update Unit. This mechanism aims to enable the knowledge memory to remember what knowledge has been used and update the memory accordingly. Another difference compared with traditional knowledge representations lies in that our proposed knowledge memory is graph-structured. Updating each memory entry will stimulate its neighboring entries and transmit their information in the update operation. Formally, in $t^{th}$ reasoning step, each memory entry is updated based on its previous memory state, its neighborhood's previous state and the current control state as:
\begin{align}
\label{eq:update}
 &m_j^{(t+1)}=\textbf{W}_{11}[\bm{m}_j^{(t)},\bm{c}_j^{nei},\bm{h}^{(t)}]\\
 &\bm{c}_j^{nei}=\sum_{k\in\Ncal_j}\textbf{W}_{12}[\bm{m}_k^{(t)},\bm{r}_{jk}]
\end{align}
where $\Ncal_i$ represents a set of 1-hop neighboring nodes regarding the memory entity $m_{j}$ and $\bm{c}_j^{nei}$ is the contextual memory representation. %This update gives each memory cell the capability to capture the graph structural information by embedding the neighboring information in its representation. 
Then the updated memory is served as the new knowledge memory used in the next reasoning step.

\subsubsection{Multimodal Feature Fusion Module}

After $T$ reasoning steps, the model collects concept-relevant knowledge for concept $v_i^F$ from the visual graph and the semantic graph independently and generate the corresponding knowledge representations denoted as $\bm{h}_{V_i}^{(T+1)}$ and $\bm{h}_{S_i}^{(T+1)}$ respectively. Then we fuse the complementary knowledge from the three modalities to form the final concept representation $\bm{v}^{F}_i$ via gate mechanism as:
\begin{align}
\label{eq:gate}
&\bm{gate}_i=\sigma(\textbf{W}_{10}[\bm{h}_{V_i}^{(T+1)}, \bm{h}_{S_i}^{(T+1)}, \hat{\bm{v}}_i^F])\\
&\widetilde{\bm{v}}^{F}_i=\textbf{W}_{11}(\bm{gate}_i\circ[\bm{h}_{V_i}^{(T+1)}, \bm{h}_{S_i}^{(T+1)}, \hat{\bm{v}}_i^F])
\end{align}
where $\sigma$ is sigmoid function and `$\circ$' is element-wise product. 

\subsection{Global Assessment and Answer Prediction}
\label{subsec:globalAnswer}

All the concepts $\{\widetilde{\bm{v}}_i^F\}_{i=1}^{\Ncal^F}$ are fed into a graph neural network \cite{hamilton2017inductive}  to globally compare with each other, which imitates the MLT in our understanding system. %The concatenation of entity representation $\hat{\bm{v}}_i^F$ and question embedding $\bm{q}$ 
The output embedding of each concept in GNN is passed to a binary classifier to predict its probability as the answer , \textit{i.e.} $\hat{y}_i=p_\theta([\widetilde{\bm{v}}_i^F,\bm{q}])$. Since there is one entity annotated as the ground-truth answer and the rest entities are all served as negative answers in each training sample,  it is necessary to use weighted binary cross-entropy loss to deal with the imbalanced training data as:
\begin{equation}
l_n=-\sum_{i\in \Ncal^F}\big[a\cdot{y}_i\ln \hat{y}_i + b\cdot(1-{y}_i)\ln(1-\hat{y}_i)\big]
\end{equation}
where ${y}_i$ is the ground truth label for $v_i^F$ and $a, b$ represent loss function weights for positive and negative samples respectively. The entity corresponding to the concept with the largest probability is selected as the final answer.

\section{Experiments}
\label{sec:experiments}

\subsection{Datasets and Evaluation Metrics} 

\textbf{FVQA:} 
The FVQA dataset \cite{wang2018fvqa} consists of  2,190 images, 5,286 questions and a knowledge base of 193,449 facts. The knowledge base is constructed by extracting the top visual concepts from all the images in the dataset and querying those concepts from three knowledge bases, including DBPedia \cite{auer2007dbpedia}, ConceptNet \cite{speer2017conceptnet} and WebChild \cite{tandon2014webchild}. For each image-question pair in the dataset, the task aims to choose an entity in a supporting fact from the knowledge base as the answer by jointly considering the given question and image.

\textbf{Visual7W+KB:} The Visual7W dataset \cite{zhu2016visual7w} is built based on a subset of images from Visual Genome \cite{krishna2017visual}, which includes questions in terms of (what, where, when, who, why, which and how) along with the corresponding answers in a multi-choice format. However, most of questions of Visual7W solely base on the image content which don't require external knowledge. Furthermore, \cite{li2017incorporating} generated a collection of knowledge-based questions based on the test images in Visual7W by filling a set of question-answer templates that need to reason on both visual content and external knowledge. We denoted this dataset as Visual7W+KB in our paper. In general, Visual7W+KB consists of 16,850 open-domain question-answer pairs based on 8,425 images in Visual7W test split. Different from FVQA, Visual7W+KB uses ConceptNet to guide the question generation but doesn't provide a task-specific knowledge base.  In our work, we also leverage ConceptNet to retrieve the supporting knowledge and select one entity as the predicted answer. 

\textbf{OK-VQA:} \cite{marino2019ok} proposed  the Outside Knowledge VQA (OK-VQA) dataset, which is the largest KVQA dataset at present. Different from existing KVQA datasets, the questions in OK-VQA are manually generated by MTurk workers, which are not derived from specific knowledge bases. Therefore, it requires the model to retrieve supporting knowledge from open-domain resources, which is much closer to the general VQA but more challenging for existing models. OK-VQA contains 14,031 images which are randomly collected from MSCOCO dataset \cite{lin2014microsoft}, using the original 80k-40k training and validation splits as train and test splits. OK-VQA contains 14,055 questions covering a variety of knowledge categories such as science \& technology, history, and sports. %10 answers are collected for each image-question pair and the most frequent answer is selected as the correct answer.
In our work, we leverage ConceptNet to retrieve the supporting knowledge and select one entity as the predicted answer.

\textbf{Evaluation Metrics:} We follow the metrics in \cite{wang2018fvqa} to evaluate the question answering performance. The top-1 and top-3 accuracy is calculated for each model. The averaged accuracy of 5 test splits is reported as the overall accuracy. %The predicted answer is correct if the string matches the corresponding ground-truth answer. All the answers have been normalized to eliminate the differences caused by singular-plurals, cases, punctuations, \textit{etc.}.  

\subsection{Implementation Details} 
For  the question representation, each question is tokenized and each word is embedded using 300-dimensional GloVe word embeddings \cite{pennington2014glove}. The maximum sentence length of question is set to 20 and questions shorter than 20 words are padded with zero vectors. The sequence of embedded words is then fed into LSTM and the dimension of the hidden layer in LSTM is set to 512. 
For constructing the semantic graph, we first generate dense captions with DenseCap \cite{johnson2016densecap}. Since some captions with low confidence are likely to introduce unexpected noise and too many captions will decrease the computation efficiency, we select top-12 dense captions with highest scores to eliminate unexpected noise caused by low confidence captions.
For the image representation, we extract 2048-dimensional object detection features, 4-dimensional spatial features, object labels with known bounding boxes from pre-trained Faster R-CNN \cite{ren2015faster} model in conjunction with ResNet-101 \cite{he2016deep}. The number of detected objects is fixed to 36. The Faster R-CNN model is trained over 1,600 selected object classes and 400 attribute classes, in a similar way as the bottom-up attention model \cite{anderson2018bottom}.

We set $a=0.7$ for positive samples and $b=0.3$ for negative samples in the binary cross-entropy loss function. 
Our model is trained by Adam optimizer with 10 epochs, where the mini-batch size is 16 and the dropout ratio is 0.5. For the strategy of learning rate, we first apply warm up strategy for 2 epoches with initial learning rate $1\times 10^{-3}$ and warm-up factor 0.2. Then we adopt cosine annealing learning strategy with initial learning rate $\eta_{max}=1\times 10^{-3}$ and termination learning rate $\eta_{min}=3.6\times10^{-4}$ for the rest epoches.

\subsection{Comparison with State-of-the-Art Methods}

\subsubsection{Experimental results on FVQA}

\begin{table}[t]
\renewcommand\arraystretch{1}
	\centering
	\setlength{\abovecaptionskip}{7pt}
	\newcommand{\tabincell}[2]{\begin{tabular}{@{}#1@{}}#2\end{tabular}}
	\resizebox{0.6\columnwidth}{!}{\scriptsize
		\begin{tabular}{l|cc}
			\hline
			\multirow{2}*{\bf Method} & \multicolumn{2}{c}{{\bf Overall Accuracy} }\\
			\cline{2-3}  & {\bf top-1} & {\bf top-3}\\
			\hline
			LSTM-Q+Image+Pre-VQA \cite{wang2018fvqa} & $24.98$ & $40.40$ \\
			
			Hie-Q+Image+Pre-VQA \cite{wang2018fvqa}& $43.14$ & $59.44$ \\
			
			FVQA (top-3-QQmaping) \cite{wang2018fvqa}& $56.91$ & $64.65$ \\
			
			FVQA (Ensemble) \cite{wang2018fvqa}& $58.76$ & - \\
			
			Straight to the Facts (STTF) \cite{narasimhan2018straight}& $62.20$ & $75.60$ \\
			
			Reading Comprehension \cite{li2019visual}& $62.96$ & $70.08$ \\
			
			Out of the Box (OB) \cite{narasimhan2018out}&$69.35$&$80.25$\\
			\hline
			Human 
			\cite{wang2018fvqa} &$77.99$&-\\
			\hline
			{\bf GRUC (ours)}&$\bm {79.63}$&$\bm{ 91.20}$\\
			
			\hline
		\end{tabular}
	}
	\caption{\small{State-of-the-art comparison on FVQA dataset.}}
	\label{table:FVQA_SOTA}
	\vspace{-2mm}
\end{table}

We compare our GRUC model with state-of-the-art models on FVQA dataset. The baseline models can be classified into three sets, including CNN-RNN based approaches, semantic parsing based approaches and learning-based approaches. The CNN-RNN based approaches \cite{wang2018fvqa} include  LSTM-Q+Image+Pre-VQA  and Hie-Q+Image+Pre-VQA. The semantic parsing based approaches \cite{wang2018fvqa} include FVQA (top-3-QQmaping) and FVQA (Ensemble). The learning based approaches include  Straight to the Facts (STTF) \cite{narasimhan2018straight}, Out of the Box (OB) \cite{narasimhan2018out}, and Reading Comprehension based approach \cite{li2019visual}.

Our model consistently outperforms all the approaches on all the metrics and achieves remarkable 10.28\% boost on top-1 accuracy and 10.95\% boost on top-3 accuracy compared with the state-of-the-art model OB \cite{narasimhan2018out}. The model OB is most relevant to GRUC in that it leverages graph convolutional networks to jointly assess all the entities in the fact graph. However, it introduces the global image features equally to all the entities without selection. By collecting question-oriented visual and semantic information via the memory-based recurrent reasoning network, our model gains remarkable improvement. It's worth to note that our model outperforms Human performance by 1.64\% on top-1 accuracy. To some extent, this results prove the effectiveness of the proposed  understanding system in \cite{McClelland2019language} since our model is designed by emulating the structure of the proposed system. In other words, the model contains distinct modules that represent each input modality and computes the representation of concepts through a mutual constraint to combine linguistic and non-linguistic inputs. Emulating this architecture in model design could contribute to achieving human-level understanding ability.

%Since OB model doesn't consider the relations between entities, 
%We also omit the relationship features in our full model and the resulting Mucko (w/o relations) model still outperforms OB by more than 3\% on average, which further proves the advantages of graph-based cross-modal reasoning compared with simple cross-modal fusion.
%These results demonstrate that visual understanding from both apperance and semantic levels and adaptively aggregating information from different modalities in cross-modal graph neural network are conducive to FVQA problem.

\subsubsection{Experimental results on Visual7W-KB}
\begin{table}[t]
\renewcommand\arraystretch{1}
	\centering
	\setlength{\abovecaptionskip}{7pt}
	\newcommand{\tabincell}[2]{\begin{tabular}{@{}#1@{}}#2\end{tabular}}
	\resizebox{0.6\columnwidth}{!}{\scriptsize
		\begin{tabular}{l|cc}
			\hline
			\multirow{2}*{\bf Method} & \multicolumn{2}{c}{{\bf Overall Accuracy} }\\
			\cline{2-3}  & {\bf top-1} & {\bf top-3}\\
			\hline
			KDMN-NoKnowledge \cite{li2017incorporating} & $45.1$ & - \\
			
			KDMN-NoMemory\cite{li2017incorporating}&$51.9$&-\\
			
			KDMN\cite{li2017incorporating}&$57.9$&-\\
			
			KDMN-Ensemble\cite{li2017incorporating}&$60.9$&-\\
			
			Out of the Box (OB) \cite{narasimhan2018out}&$57.32$&$71.61$\\
			\hline
    
			{\bf GRUC (ours)}&$\bm {69.03}$&$\bm{ 88.12}$\\
			
			\hline
		\end{tabular}
	}
	\caption{\small{State-of-the-art comparison on Visual7W+KB dataset.}}
	\label{table:V7W_SOTA}
	
\end{table}

The comparison of state-of-the-art models on Visual7W-KB dataset is shown in Table \ref{table:V7W_SOTA}. The compared baselines contains two sets, i.e. memory-based approaches and a graph-based approach. The memory-based approaches \cite{li2017incorporating} include KDMN-NoKnowledge (w/o external knowledge), KDMN-NoMemory (attention-based knowledge incorporation), KDMN (dynamic memory network based knowledge incorporation) and KDMN-Ensemble (several KDMN models based ensemble model). We also test the performance of Out of the Box (OB) \cite{narasimhan2018out} on Visual7W-KB and report the results in Table \ref{table:V7W_SOTA}.

As consistent with the results on FVQA, we achieve a significant improvement (8.13\% on top-1 accuracy and 16.51\% on top-3 accuracy ) over state-of-the-art models. Note that our proposed GRUC network is an single-model, which outperforms the existing ensembled model \cite{li2017incorporating}. We believe that the performance can be further improved if the technique of ensemble is involved in our model. 

\subsubsection{Experimental results on OK-VQA}
\begin{table*}
\renewcommand\arraystretch{1}
\centering
\setlength{\abovecaptionskip}{7pt}
\resizebox{\textwidth}{!}{\scriptsize
\begin{tabular}{c|cc|ccccccccccc}
\hline
\multirow{2}*{\bf Method} & \multicolumn{2}{c|}{{\bf Overall Accuracy} }\\
\cline{2-3}   & {\bf top-1} & {\bf top-3}& VT & BCP & OMC & SR & CF & GHLC & PEL & PA & ST & WC & Other\\
\hline
Q-Only \cite{marino2019ok}& 14.93 &-& 14.64 & 14.19 & 11.78 & 15.94 & 16.92 & 11.91 & 14.02 & 14.28 & 19.76 & 25.74 & 13.51 \\
MLP \cite{marino2019ok}& 20.67 &-& 21.33 & 15.81 & 17.76 & 24.69 & 21.81 & 11.91 & 17.15 & 21.33 & 19.29 & 29.92 & 19.81 \\
BAN \cite{kim2018bilinear} & 25.17&- & 23.79 & 17.67 & 22.43 & 30.58 & 27.90 & 25.96 & 20.33 & 25.60 & 20.95 & \textbf{40.16} & 22.46 \\
MUTAN \cite{ben2017mutan} & 26.41&- & 25.36 & 18.95 & 24.02 & 33.23 & 27.73 & 17.59 & 20.09 & 30.44 & 20.48 & 39.38 & 22.46 \\
ArticleNet (AN) \cite{marino2019ok}& 5.28&- & 4.48 & 0.93 & 5.09 & 5.11 & 5.69 & 6.24 & 3.13 & 6.95 & 5.00 & 9.92 & 5.33 \\
BAN + AN \cite{marino2019ok}& 25.61 & -&24.45 & 19.88 & 21.59 & 30.79 & 29.12 & 20.57 & 21.54 & 26.42 & 27.14 & 38.29 & 22.16 \\
MUTAN + AN\cite{marino2019ok} & 27.84 & -& 25.56 & 23.95& 26.87 & \textbf{33.44} & 29.94 & 20.71 & 25.05 & 29.70 & 24.76 & 39.84 & 23.62  \\
BAN/AN oracle \cite{marino2019ok} & 27.59  & -& 26.35 & 18.26 & 24.35 & 33.12 & 30.46 & \textbf{28.51} & 21.54 & 28.79 & 24.52 & 41.4 & 25.07 \\
MUTAN/AN oracle \cite{marino2019ok} & 28.47  & -& 27.28 & 19.53 & 25.28 & 35.13 & \textbf{30.53} & 21.56 & 21.68 & \textbf{32.16} & 24.76 & 41.4 & 24.85 \\ 
\hline
\textbf{GRUC (ours)} & \textbf{29.87} & \textbf{32.65}&\textbf{29.84} & \textbf{25.23} & \textbf{30.61} &30.92 & 28.01 & 26.24 & \textbf{29.21} & 31.27 & \textbf{27.85} & 38.01 & \textbf{26.21}  \\\hline
\end{tabular}
}
\caption{\small{State-of-the-art comparison on OK-VQA dataset. We show the results for the full OK-VQA dataset and for each knowledge category (top-1 accuracy): Vehicles and Transportation (VT); Brands, Companies and Products (BCP); Objects, Material and Clothing (OMC); Sports and Recreation (SR); Cooking and Food (CF); Geography, History, Language and Culture (GHLC); People and Everyday Life (PEL); Plants and Animals (PA); Science and Technology (ST); Weather and Climate (WC); and Other.}}
\label{table:OK-VQA_SOTA}
\vspace{-2mm}
\end{table*}

We also report the quantitative performance on the challenging OK-VQA dataset in Table \ref{table:OK-VQA_SOTA}. We compare our model with three kinds of existing models, including current state-of-the-art VQA models, knowledge-based VQA models and ensemble models. The VQA models contain Q-Only \cite{marino2019ok}, MLP \cite{marino2019ok}, BAN \cite{kim2018bilinear}, MUTAN\cite{kim2018bilinear}. The knowledge-based VQA models \cite{marino2019ok} consist of ArticleNet (AN), BAN+AN and MUTAN+AN. The ensemble models \cite{marino2019ok}, i.e. BAN/AN oracle and MUTAN/AN oracle,  simply take the raw ArticleNet and VQA model predictions, taking the best answer (comparing to ground truth) from either. We report the overall performance (top-1 and top-3 accuracy) as well as breakdowns for each of the knowledge categories (top-1 accuracy). We have the following two observations from the results:

First, our model consistently outperforms all the compared models on the overall performance. Even the state-of-the-art models (BAN and MUTAN) specifically designed for VQA tasks, they get inferior results compared with ours. This indicates that general VQA task like OK-VQA cannot be simply solved by a well-designed model, but requires the ability to incorporate external knowledge in an effective way. Moreover, our model outperforms knowledge-based VQA models including both single models (BAN+AN and MUTAN+AN) and ensemble models (BAN/AN oracle and MUTAN/AN oracle), which further proves the advantages of our knowledge incorporating mechanism based on both multimodal knowledge graphs and memory-enhanced recurrent reasoning network. 

Second, the improvement of our model on OK-VQA is not that remarkable compared to the performance on FVQA and Visual7W-KB. We believe that this phenomenon is mostly due to the following two reasons: (1) Questions in the OK-VQA dataset are more diverse and complex, which is more challenging for machines to understand accurately. The questions in FVQA and Visual7W-KB are generated when given the images and supporting facts upon the pre-defined templates or relations. This mechanism constrains the answers in a specific knowledge base and guides the model to operate in a reverse way of the question generation process to predict the correct answers with high probability. On the contrary, questions in OK-VQA are totally free-form ones that generated by MTurk workers and thus containing more unique questions and words with less bias compared with other datasets. This increases the difficulty to understand the questions accurately. (2) OK-VQA requires a wide range of knowledge beyond a specific knowledge base. Looking at the category breakdowns in Table \ref{table:OK-VQA_SOTA}, baseline models achieve relatively high performance for SR, CF, GHLC, PA and WC categories while our model performs better for the remaining categories. Since the baseline models refer to the Wikipedia while our model refers to ConceptNet, the performance in the category breakdowns perhaps suggests that each knowledge base just provides a portion of required knowledge. To improve the overall performance on OK-VQA, we should better  comprehensively consider knowledge bases that cover commonsense, visual knowledge, Wikipedia knowledge and even professional knowledge.

\subsection{Ablation Study}
\begin{table*}
\renewcommand\arraystretch{1}
	\centering
	\setlength{\abovecaptionskip}{7pt}
	\resizebox{\textwidth}{!}{\scriptsize
		\begin{tabular}{l|l|cc|cc|cc}
			\hline
			\multicolumn{2}{l|}{\multirow{2}*{\bf Method}}&
			\multicolumn{2}{c|}{{\bf FVQA} }&
			\multicolumn{2}{c|}{{\bf Visual7W+KB} }&
			\multicolumn{2}{c}{{\bf OK-VQA} }
			\\
			\cline{3-8}
			\multicolumn{2}{l|}{~}  & {\bf top-1} & {\bf top-3}& {\bf top-1} & {\bf top-3}&{\bf top-1} & {\bf top-3}
			\\
			\hline
			\multicolumn{2}{l|}{ {\bf GRUC} (full model)}& ${\bf79.63}$ & ${\bf91.20}$ & ${\bf69.03}$ & ${\bf88.12}$& $\bm {29.87}$&$\bm{32.65}$\\
			\hline
			1& w/o Semantic Graph & $78.05$ & $87.70$ & $67.01$ & $84.91$& $28.30$ & $31.02$\\
			2&w/o Visual Graph& $76.98$ & $83.15$& $66.38$ & $79.80$ & $28.02$ & $29.52$\\
			3& w/o Semantic Graph \& Visual Graph & $20.43$ &$29.10$& $17.88$ & $28.43$& $12.11$ & $13.96$ \\
			\hline
			4& w/o Neighbor Aggregation (Control Unit) & $78.53$ & $89.34$ & $68.34$ & $85.67$& $28.20$ & $30.89$\\
			5& w/o Neighbor Aggregation (Update Unit) &$77.61$ & $88.05$ & $66.52$ & $82.04$& $25.74$ & $27.62$\\
			6& w/o GRUC Module 
			&$70.87$&$78.70$& $57.22$ & $70.80$& $18.65$ & $20.91$\\
			\hline
			7& w/o Intra-Modal Knowledge Selection &$74.85$&$80.63$& $67.28$ & $85.41$& $26.49$ & $27.56$\\
			\hline
			8& w/o Global Assessment &$79.10$&$90.54$& $68.43$ & $87.69$& $29.80$ & $32.11$\\
			\hline
		\end{tabular}
	}
	\caption{\small{Ablation study of key components on FVQA, Visual7W-KB and OK-VQA.}}
	\label{table:ablation}
	\vspace{-2mm}
\end{table*}
Since our model contains multiple essential components, we test a series of variations on the three benchmark datasets to verify the influence of each component. The experimental results are shown in Table \ref{table:ablation}.
%\textit{Visual}, \textit{Semantic} and \textit{Fact} denote information from visual graph, semantic graph and fact graph, respectively. \textit{V-to-F Conv.} and \textit{S-to-F Conv.} denote visual-to-fact convolution and semantic-to-fact convolution.
%\textit{V-to-F Concat.} denotes gathering information from visual graph to fact graph by concatination of the entitiy embedding and the averaged object embeddings while %\textit{S-to-F Concat.} denotes gathering information from semantic graph to fact graph by concatination of the entitiy embedding and the averaged caption embeddings.
%\textit{V-to-F Concat.} and \textit{S-to-F Concat.} denote gathering information from visual and semantic to fact graph by concating averaged region embeddings and caption embeddings with entities.
%\textit{\#Step} denotes the number of reasoning steps. 
%We evaluate 7 variant versions and conclude with three observations:
% \textbf{(1) The influence of each layer of graphs.} 
\subsubsection{Influence of Different Knowledge Modalities}
As demonstrated in Section \ref{sec:intro}, we believe that different modalities can provide complementary knowledge for answer inference. In this Section, we conduct ablation study to prove the indispensable role of each modality by the following variations:

\begin{itemize}
\item[-] \textbf{w/o Semantic Graph (model `1')}: this model removes the semantic graph in graph construction and the follow-up reasoning process.

\item[-] \textbf{w/o Visual Graph (model `2')}: this model removes the visual graph in graph construction and the follow-up reasoning process.

\item[-] \textbf{w/o Semantic Graph \& Visual Graph (model `3')}: this model simultaneously removes the visual graph and the semantic graph in graph construction and the follow-up reasoning process.

\end{itemize}

The experimental results are shown in the first block in Table \ref{table:ablation}. We observe that the top-1 and top-3 accuracy of `1' and `2' all decrease compared with the full model on the three datasets, which indicates that both semantic and visual graphs are beneficial to 
provide valuable evidence for answer inference. Thereinto, the visual information has greater impact than the semantic part, proving that the image content still plays essential role in KVQA tasks. When removed both semantic and visual graphs, `3' results in a significant decrease. It gives us insight that factual knowledge only is entirely insufficient to answer the question. By incorporating three of the knowledge modalities, we achieve the best performance.  

\subsubsection{Influence of the GRUC Module}

As the key component in our model, the GRUC module has two advantages over the existing reasoning models: first, it considers the structure information in the knowledge base and involves the structures in the reasoning process; second, the multi-step reasoning process via recurrent GRUC modules achieves greater reasoning ability compared with the state-of-the-art OB model \cite{narasimhan2018out} merely applying feature concatenation. We justify these two advantages by the following variations:

\begin{itemize}
\item[-] \textbf{w/o Neighbor Aggregation (Control Unit) (model `4')}: this model removes the neighborhood information, i.e. $\bm{c}^F_i$ in Equation \ref{eq:controInit}, in initializing the control unit.

\item[-] \textbf{w/o Neighbor Aggregation (Update Unit) (model `5')}: this model removes the neighborhood information, i.e. $\bm{c}_j^{nei}$ in Equation \ref{eq:update}, in the update unit.

\item[-] \textbf{w/o GRUC network (model `6')}: this model replaces the GRUC network and multimodal feature fusion introduced in Section \ref{subsec:crossReasoning} by direct concatenation, \textit{i.e.} concatenating the mean pooling of all the semantic/visual node features with each entity feature.

\end{itemize}

The experimental results are shown in the second block in Table \ref{table:ablation}. The performance on three datasets decreases slightly when remove the neighborhood information from either the control unit or the update unit. It indicates that preserving the structural information when incorporating the knowledge brings richer semantics for answer prediction. The performance decreases more than 10\% when replacing the GRUC network by simple concatenation, which proves the advantages of the proposed recurrent reasoning process in gathering complementary evidence from different modalities.

\subsubsection{Influence of the Intra-Modal Knowledge Selection}

The first stage in our model is to select knowledge from each modalities under the guidance of the question independently. Since most questions are referring to a small portion of knowledge, this stage aims to choose relevant knowledge and avoid unexpected noise for improving the performance. We prove this motivation by the variation below:   

\begin{itemize}
\item[-] \textbf{w/o Intra-Modal Knowledge Selection (model `7')}: this model removes the intro-modal knowledge selection process introduced in Section \ref{subsec:IntraConv}.
\end{itemize}

The experimental results are shown in the third block in Table \ref{table:ablation}. We observe that all the metrics decrease remarkably on all the three datasets. It indicates that `intuitive' knowledge filtering before `clever' knowledge reasoning is effective to bring extra improvement. 

\subsubsection{Influence of the Global Assessment}

The last stage in our model is to globally assess all the concepts via GNN and choose the optimal one as the answer. To prove the influence of this process, we further conduct the following ablation study: 

\begin{itemize}
\item[-] \textbf{w/o Global Assessment (model `8')}: this model removes the GNN operated on all the concepts in Section \ref{subsec:globalAnswer} and feeds the embedding of each concept in $\{\widetilde{\bm{v}}_i^F\}_{i=1}^{\Ncal^F}$ directly to the binary classifier.
\end{itemize}

The experimental results are shown in the last block in Table \ref{table:ablation}. The performance consistently decreases on all the datasets. However, the decrease is relatively smaller compared with other models. We think that the GNN model used in this process is relatively simple and more effective global assessment approach perhaps can bring more improvement.

\begin{figure*}[htpb]
	\centering
	\setlength{\abovecaptionskip}{7pt}
	\includegraphics[width=\textwidth]{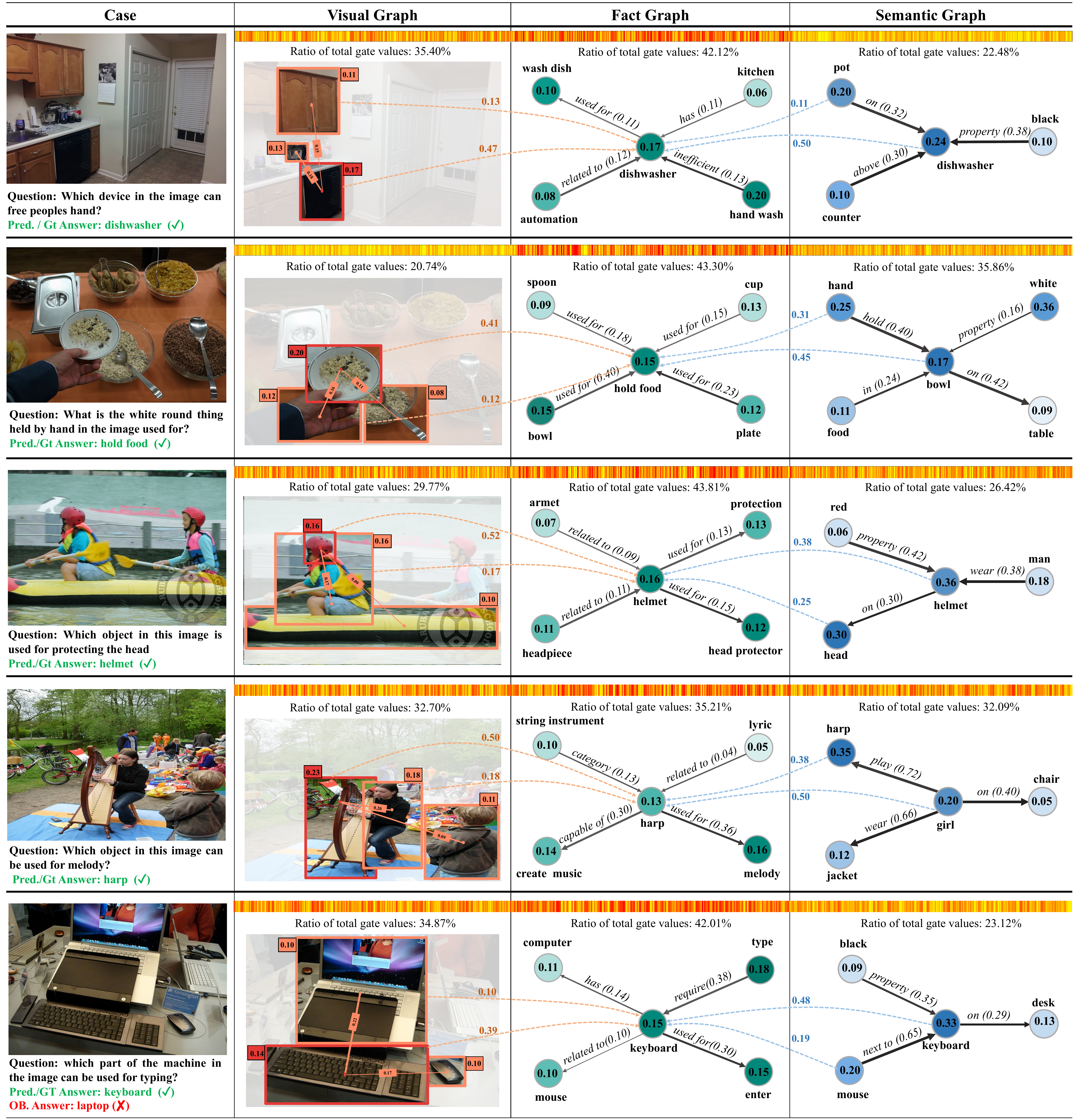}
	\caption{\small{Visualization for our model. Visual graph highlights the most relevant subject (red box) according to attention weights of each object ($\alpha^V$ in Eq. \ref{eq:node attention}) and the objects (orange boxes) with top-2 attended relationships ($\beta^V$ in Eq. \ref{edge attention}). Fact graph shows the predicted entity (center node) and its top-4 attended neighbors ($\alpha^F$ in Eq. \ref{eq:node attention}). Semantic graph shows the most relevant concept (center node) and its up to top-4 attended neighbors ($\alpha^S$ in Eq. \ref{eq:node attention}). Each edge is marked with attention value ($\beta^{F/S}$ in Eq. \ref{edge attention}). Dash lines represent memory read attention weights at reasoning step $T$ ($\gamma^{(T)}$ in Eq. \ref{mem_att}).
			%, where orange lines indicate top-2 visual-to-fact conv. weights ($\gamma^{\textit{\text{V-F}}}_{ij}$ in Eq. \ref{eq:VtoF}) and blue lines indicate top-2 semantic-to-fact conv. weights ($\gamma^{\textit{\text{S-F}}}_{ij}$ in Eq. \ref{eq:VtoF}).
			%The heatbar denotes the gate value and reveals the amount of knowledge derived from each module computed from Eq.\ref{eq:gate}. for inferring the answer.
			The thermogram on the top visualizes the gate values ($\bm{gate}_i$ in Eq. \ref{eq:gate}) of visual embedding (left), entity embedding (middle) and semantic embedding (right). 
			%The ratio of gate values for the visual, semantic and fact graph is computed from Eq. \ref{eq:gate}.
	}} 
	\label{fig:case}
\end{figure*} 

\begin{figure*}[tpb]
\setlength{\abovecaptionskip}{7pt}
	\includegraphics[width =\textwidth]{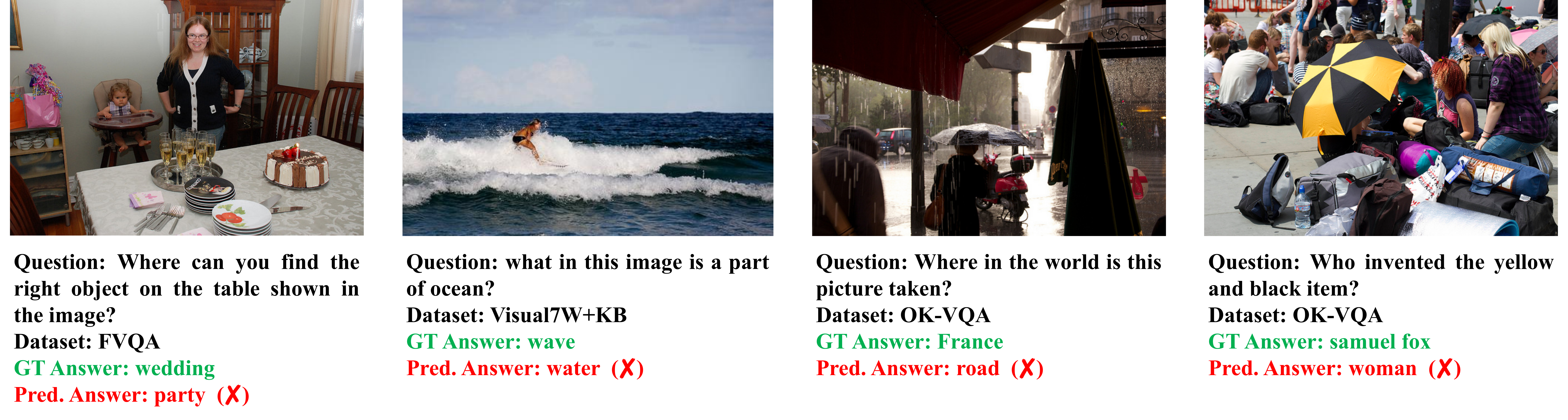}
	\caption{ Visualization of four failure cases.
	}
	\label{fig:failure}
	\vspace{-2mm}
\end{figure*}

\subsection{Interpretability}

Our model is interpretable by visualizing the attention weights and gate values in the cross-modal heterogeneous graph reasoning process.
From case study in Figure \ref{fig:case} and Figure \ref{fig:failure}, we conclude with the following three insights: 
%Due to space limitation, we only visualize partial nodes in three graphs. 

\textbf{GRUC is capable to reveal the knowledge selection mode from different modalities.} In Figure \ref{fig:case}, the first four examples indicate that GRUC captures the most relevant visual, semantic and factual evidence (according to intra-modal attention weights) as well as complementary information across modalities (according to attention weights in the Read Unit of the GRUC module). The \textit{ratio of total gate values} reveals the amount of information derived from each graph. We denote the sum of all the gate values for the visual dimensions, fact dimensions and semantic dimensions in Equation \ref{eq:gate} as $G_v$, $G_f$ and $G_s$, respectively. We denote the sum of all the gate values for all the dimensions in Equation \ref{eq:gate} as $G$. The \textit{ratio of total gate values} for the visual graph, fact graph and semantic graph is defined as $\frac{G_v}{G}$, $\frac{G_f}{G}$ and $\frac{G_s}{G}$, respectively. In most cases, factual knowledge provides predominant clues compared with other modalities from the observation of the ratio of gate values. It is because that KVQA tasks rely on external knowledge to a great extent. Furthermore, more evidence comes from the semantic modality when the question involves complex relationships. % like the second example compared with the first one. 
For instance, the question in the first case asking about single object `devise' requires more visual information while the question in the second case involving the relationship between `hand' and `while round thing' needs more semantic clues.

\textbf{GRUC has advantages over the state-of-the-art model.} In Figure \ref{fig:case}, the fifth example compares the predicted answer of Out of the Box (OB) \cite{narasimhan2018out} with GRUC. GRUC collects relevant visual and semantic evidence to make each entity discriminative enough for predicting the correct answer while OB failing to distinguish representations of `laptop' and `keyboard' without feature selection.

\textbf{GRUC fails mostly in three conditions: highly relevant answers, inadequate visual evidence and limited external knowledge.} Figure \ref{fig:case} shows four failure cases. (1) Some cases fail when the predicted answer is quite relevant to the ground truth answer. In the first case, it's reasonable that both `wedding' and `party' may have cakes. There is no further evidence from the image to decide which situation is more accurate. We can explain the second case by the similar reason. (2) Some cases fail when there is no enough evidence to infer the correct answer, such as the third sample in Figure \ref{fig:case}. (3) Some other failure cases are due to the lack of required knowledge in the provided knowledge base. In the last case in Figure \ref{fig:case}, there is no fact about `samuel fox' in ConceptNet. Therefore, comprehensively considering multiple knowledge bases to cover a wider range of knowledge is important to improve the KVQA ability.

\begin{figure}[tpb]
	\centering
	\setlength{\abovecaptionskip}{7pt}
	\includegraphics[width=0.7\columnwidth]{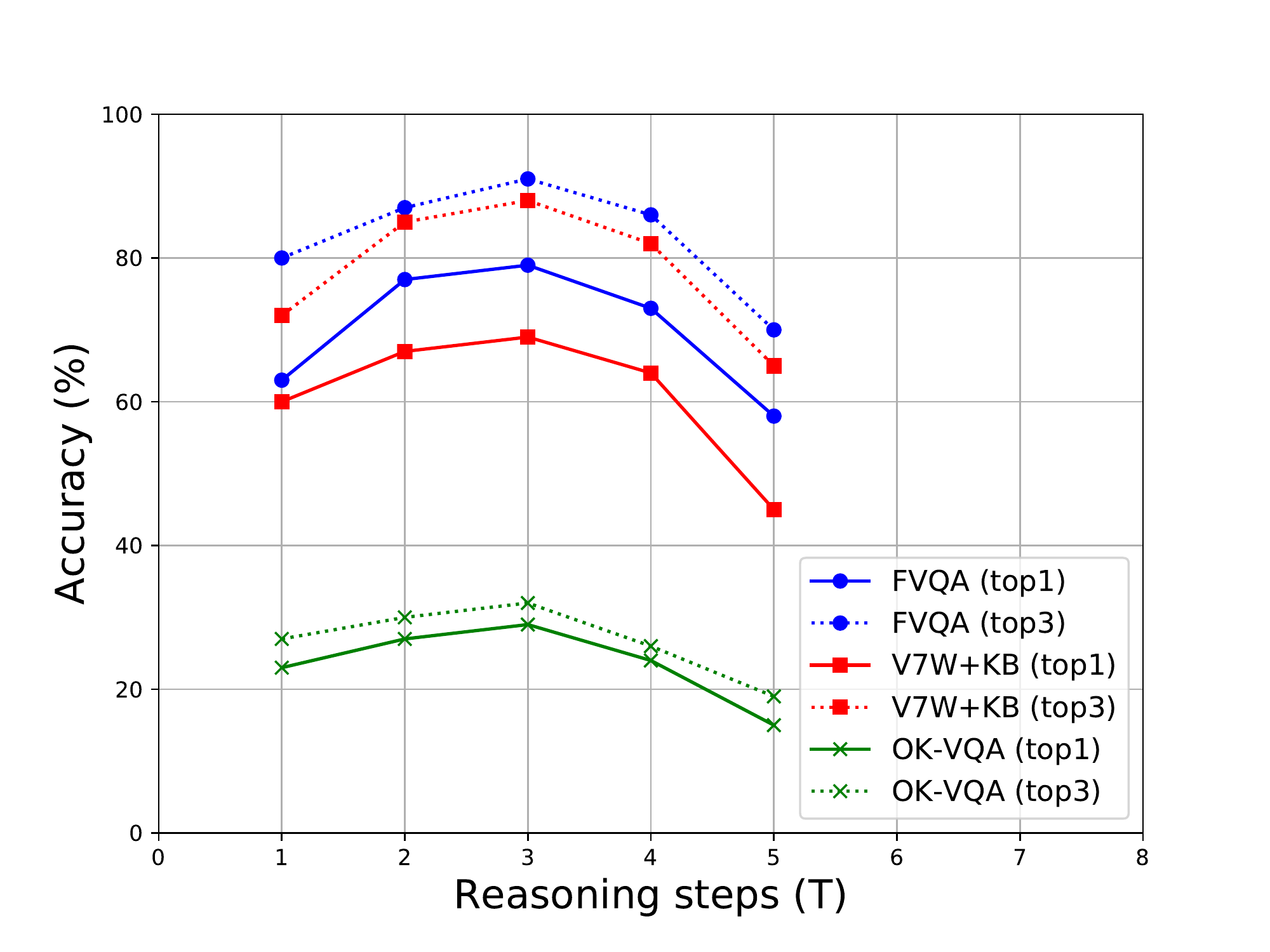}
	\caption{\small{Overall accuracy with different number of reasoning steps in the GRUC network. The blue solid and dotted lines respectively denote the top-1 and top-3 accuracy on FVQA. The red solid and dotted lines respectively denote the top-1 and top-3 accuracy on Visual7W+KB. The green solid and dotted lines respectively denote the top-1 and top-3 accuracy on Ok-VQA.}}
	\label{fig:step}
\end{figure} 

\begin{table}[t]
	\centering
	\setlength{\abovecaptionskip}{7pt}
	\resizebox{0.5\textwidth}{!}{\scriptsize
	\begin{tabular}{l|ccc}
		\hline
		{\bf  \#top-k dense captions }  &5& 10 & 20\\
		\hline
		{\bf top-1 accuracy}&70.20 & {\bf 73.06} & 65.40 \\
		{\bf top-3 accuracy} & 82.65 & {\bf85.94} & 75.98 \\
		\hline
	\end{tabular}
		}
	\caption{Overall accuracy with different number of dense captions.}
	\label{table:captionNum}
	\vspace{-2mm}
\end{table}

\subsection{Parameter Analysis}
We evaluate the influence of different number of dense captions in Table \ref{table:captionNum}. The results show that 10 captions achieves the best performance on both top-1 and top-3 accuracy. We further evaluate the influence of different number of reasoning steps $T$ in the GRUC network. Figure \ref{fig:step} shows the top-1 and top-3 accuracy  on FVQA, Visual7W+KB and OK-VQA when setting the number of reasoning steps in the range of 1 to 5.
%i.e. conducting cross-modal heterogeneous graph reasoning $T$ times.
We find that 3 reasoning steps achieve the best performance on the all datasets. If the number of reasoning steps less than 3, the GRUC network cannot extract adequate knowledge from each memory to support the global assessment. In contrast, too many reasoning steps may lead to over-smoothing, leading to the features of nodes converging to the similar values. Therefore, we use this setting in our full model and the ablation models.

%Our fact retrieval module achieves comparable accuracy with the state-of-the-art model ``Out-of-the-Box (OB)'' \cite{narasimhan2018out}. For fair comparison, we follow the settings in OB to retrieve top-100 relevant facts filtered by top-3 relation types as the candidates. To notice that our model achieves 3.71\% improvement over OB on downstream accuracy even with slightly inferior fact recall, which highlights the effectiveness of our cross-modal knowledge reasoning module. 

%We find that the downstream accuracy is highest when we retrieve top-100 relevant facts with top-3 relation types. 

\section{Conclusion}
\label{sec:conclution}
In this paper, we propose a graph-based recurrent reasoning network GRUC for visual question answering requiring external knowledge, which focuses on cross-modal knowledge reasoning upon graph-structured multimodal knowledge representations. We novelly depict multimodal knowledge sources by  multiple  knowledge  graphs  from  the  visual,  semantic  and factual views. The representations of different modalities are unified in graph domain, thus benefiting for relational reasoning across modalities. Meanwhile, introducing the semantic graph for high-level abstraction brings remarkable improvement in KVQA, which has been less studied in previous work. On top of these representations, we propose a new recurrent reasoning model and each  reasoning  step  is performed by  a Graph-based Read, Update, and Control (GRUC) module that conducts parallel reasoning over both visual and semantic information. GRUC is a parallel reasoning module that applies modality-oriented controllers for reasoning over different modalities in a parallel mode, which can be easily extended to involve more modalities.
Our model consistently outperforms the state-of-the-art approaches remarkably on FVQA, Visual7W-KB and OK-VQA datasets. Furthermore, the model has good interpretability of revealing the knowledge selection mode from different modalities by comprehensive visualization.
However, our model has inferior performance when open-domain knowledge is required. How to adaptively incorporate diverse knowledge bases that covering commonsense, Wikipedia knowledge and even professional knowledge for KVQA tasks will be our future work.

%\section*{References}

\section*{Acknowledgement}

This work was supported by the National Key Research and Development Program of China under Grant No.2016YFB0801003.

\bibliography{mybibfile}

\small{
~\\
\noindent\textbf{Jing Yu}  is an Assistant Professor in the Institute of Information Engineering, Chinese Academy of Sciences. She received her B.S. degree in Automation Science from Minzu University, China, in 2011, and got her M.S. degree in Pattern Recognition from Beihang University, China in 2014. She recieved her Ph.D. degree in the University of Chinese Academy of Sciences, China, in 2019. She works on Vision and Language problems, including visual question answering, visual dialogue, cross-modal information retrieval, etc.\\

\noindent\textbf{Zihao Zhu} received his B.S. degree in Computer Science and Technology from China University of Mining and Technology in 2018. Now he is a graduate student in Institute of Information Engineering, Chinese Academy of Sciences, China and School of Cyber Security, University of Chinese Academy of Sciences, China. His research interests include multimodal machine learning, visual question answering.\\

\noindent\textbf{Yujing Wang} is a senior researcher in Microsoft Research Asia. She received her B.S. and M.S. degrees from Peking University, China, in 2010 and 2013, respectively. Her current research interests include Automated Machine Learning, Natural Language Understanding, and Graph Mining.\\

\noindent\textbf{Weifeng Zhang} received his B.S degree in Electronic Information Engineering from Beijing University of Technology in 2009, got his M.S degree in Pattern Recognition from Beihang University in 2012, and got his Ph.D. degree in Computer Science from Hangzhou Dianzi University in 2019. Now, he is an associate professor in Jiaxing University. His research interests include machine learning, multimedia modeling.\\

\noindent\textbf{Yue Hu} is a Professor in the Institute of Information Engineering, Chinese Academy of Sciences. She received her Ph.D. in computer science from  University of science and technology , Beijing in 2000. She is currently a researcher in Institute of Information Engineering , Chinese Academy of Sciences . Her research interest are in the area of natural language processing and social network analysis.\\

\noindent\textbf{Jianlong Tan} received his Ph.D. degree in Institute of Computing Technology, Chinese Academy of Sciences, Beijing, in 2003. He received his B.S. and M.S. degree in Xiangtan University, China, in 1997 and 2000, respectively. He is currently a professor in Institute of Information Engineering, Chinese Academy of Sciences. His research interests are in the area of multimedia analysis and hardware algorithm design.
}
\end{document}